# Time Series of Non-Additive Metrics: Identification and Interpretation of Contributing Factors of Variance by Linear Decomposition

**Alex Glushkovsky**
BMO Financial Group

## Abstract

The research paper addresses linear decomposition of time series of non-additive metrics that allows for the identification and interpretation of contributing factors (input features) of variance. Non-additive metrics, such as ratios, are widely used in a variety of domains. It commonly requires preceding aggregations of underlying variables that are used to calculate the metric of interest. The latest poses a dimensionality challenge when the input features and underlying variables are formed as two-dimensional arrays along elements, such as account or customer identifications, and time points. It rules out direct modeling of the time series of a non-additive metric as a function of input features. The article discusses a five-step approach: (1) segmentations of input features and the underlying variables of the metric that are supported by unsupervised autoencoders, (2) univariate or joint fittings of the metric by the aggregated input features on the segmented domains, (3) transformations of pre-screened input features according to the fitted models, (4) aggregation of the transformed features as time series, and (5) modelling of the metric time series as a sum of constrained linear effects of the aggregated features. Alternatively, approximation by numerical differentiation has been considered to linearize the metric. It allows for element level univariate or joint modeling of step (2). The process of these analytical steps allows for a backward-looking explanatory decomposition of the metric as a sum of time series of the survived input features. The paper includes a synthetic example that studies loss-to-balance monthly rates of a hypothetical retail credit portfolio. To validate that no latent factors other than the survived input features have significant impacts on the metric, Statistical Process Control has been introduced for the residual time series.

## 1. Introduction

A variety of domains, including business management, rely on metrics (also known as Key Performance Indicators). Metrics time series are major business entities that appear in reports, dashboards, and presentations.

Observing volatility of the metric time series raises immediate questions:

- What are the contributing factors of the metrics' period-to-period changes?
- What is the magnitude, timing, and direction of their impacts?
- Can we study the historical values of the metric and input features to forecast future time series?
- What could be changed to bring the metric to the desired level?
- What will be the potential effects given the expected or stressed levels of the input features?

Today, there is unprecedented data collection that should supposedly help answer these questions. However, even relying on big data support, the answers are usually not trivial reporting exercises unless a metric represents an additive measure with linear effects from known contributing factors (input features).

The paper addresses cases where the metric of interest is not an additive measure (for example, a ratio), aggregations are required to get a confident estimation of the metric, and there are non-linear effects of input features that impact the metric. For instance, considering a retail credit portfolio, one of the most important business metrics is the loss-to-balance ratio (loss rate) and it is a non-additive measure. The loss amounts on an account level are quite sparse and, therefore, the calculated loss-to-balance ratio requires attention with respect to its confidence.

From another side, the number of input features, interactions between them, different natures, timings of impacts (often synchronized or intersected), *a priory* unknown characters of the dynamic responses, and limited monitoring periods - all create a very challenging analytical problem addressing questions mentioned above (Glushkovsky, 2018).

Most of the analytical approaches focusing on these questions can be split into two distinct groups: (a) modeling on the most granular element levels, such as transactions, accounts, or customer identifications, and (b) time series analysis on entity levels, such as business, portfolios, or segments.

Addressing risk factors on a customer level, the stochastic models that target credit defaults have been developed in terms of observable and unobservable risk factors (latent variables). It decomposes risk factors by their sources: customer specific (both static and dynamic), portfolio specific, macroeconomic, and market (Bucay and Rosen, 2001; Heitfield, 2005; Kupiec, 2007; Lamb and Perraudin, 2008; Hamerle *et al*, 2011).

Interpretable time series forecasting has been considered by using "seasonal-trend" decompositions (Cleveland et al., 1990) with recent advances of applying deep learning N-BEATS architectures (Oreshkin *et al*, 2020). These approaches are developed for univariate time series.

Portfolio level time series analyses are widely described by including macroeconomic variables, autocorrelation, and lagged effects (Harvey, 1989; Schmidt and Schmieder, 2007; Lamb and Perraudin, 2008; Andersson, 2011; Assouan, 2012). To analyze period-to-period changes on a segment level, Markov chains for transition matrixes have been applied (Thomas, 2009; Andersson, 2011).

Element level models are usually at the highest resolution with the possibility of aggregation to entity levels, such as portfolio or segment. However, some important metrics are not additive and are not directly applicable on an element level, such as a loss rate or loss-to-revenue ratio. Furthermore, business-oriented models on an element level are mostly point-in-time (PIT) oriented with outcomes that reflect current conditions (Heitfield, 2005). Even including macroeconomic through-the-cycle (TTC) factors as input features in such models, these factors are usually rejected and substituted by element specific variables with higher predictive power. It means that additional steps are required to fit the results of element level models to the observed metric time series. These steps are very challenging considering the non-additivity of the metrics and non-linearity of input feature effects.

Models based on time related decompositions, such as the Age-Period-Cohort (Yang and Land, 2013), focus on performance of cohorts along their age in an unstable environment. Thus, the dual-time dynamics approach has been developed and successfully applied to different problems in customer behavior, including effective stress-testing (Breeden, 2007; Breeden *et al*, 2008). It provides a distinct decomposition of vintage behaviour by independent functions of age, calendar, and quality of vintage. In this approach, some form of relationships between functions should be assumed and it requires an iterative fitting process or a non-linear optimizer to find a solution. Providing input features collected at the time of acquisition (such as risk level, product attributes, loan-to-value for secured credits, etc.), the dual-time dynamics approach is applied independently to each segment, that has been split by these input features, with the following rerun collapsing some of them and adding constraints. The macroeconomic, managerial, and seasonality factors are introduced after exogenous curve extraction (Breeden *et al*, 2008). Considering business applications, such as retail portfolios, these factors can be idiosyncratic. For example, policy changes can affect only a subgroup of customers, and be spread out over time. Having these properties of the macroeconomic, regulatory, or managerial impacts, it may significantly complicate the decomposition process.

Applying Unobserved Components (UC) models, the econometric time series can be decomposed by seasonality with trend and cycle interactions (Koopman and Lee, 2009). However, the challenging issue of this approach is the incorporation of element level inputs.

Interpretability of the modeled results is a decisive aspect in addressing the questions above. Essentially, it requires additive decomposition of the metric time series by contributing factors. Generalized Additive Models (GAMs) (Hastie and Tibshirani, 1986) or Neural Additive Models (NAMs) (Agarwal *et al*, 2021) support the interpretability and nonlinearity of relationships between input features and the target variable. Moreover, the NAMs provide joint learning of input features that can flexibly fit complex relationships while achieving state-of-the-art accuracy. However, when considering the non-additive character of the metric, GAMs or NAMs approaches cannot be applied directly. The same challenges deter straightforward implementations of LIME (Ribeiro, 2016) or Shapley (Shapley, 1953; Lundberg and Lee Su-In, 2017) techniques that allow for local additive decompositions.

To address the above-mentioned questions and challenges, linear decomposition of time series of the non-additive metrics by transformed element level input features is considered in this paper. It allows for the identification and interpretation of contributing factors of period-to-period variances of the metric time series of interest.

## 2. The Problem Definition

Let us consider a time series of the target metric $Z(t)$ that is calculated as function $F(.)$ based on $M$ underlying variables $Y_{1:M}$. Underlying variables are defined on an element level $i$, such as account or customer identifications, at discrete time points $t=1:T$:

$$Z(t)=F(\widehat{Y_1}(t), \widehat{Y_2}(t), \ldots \widehat{Y_M}(t)) \quad (1)$$

where $\widehat{Y_m}(t) = Agg_{i \in P(t)}(Y_m(i,t))$, $m=1:M$ and $Agg(.)$ is an operator of aggregation, such as count, mean, min, max, or sum; $P(t)$ is a set of elements $i$ at time $t$.

Commonly, in real-world business applications, the calculation (1) has the following two properties:

- function F(.) is non-additive, i.e., it is nonlinear in terms of $\widehat{Y_{1:M}}(t)$.
- it requires preceding aggregations of underlying variables to a level above $i$ to obtain a robust measure of $Z$, including the elimination of divisions by zero and providing sufficient sizes of aggregations.

The paper assumes that both properties are present.

Metric $Z$ describes a business characteristic of an entity, such as a business unit or a portfolio, that consists of $N(t)$ number of elements $i$ at time $t$, and that this number may vary over time reflecting changes due to acquisitions or attritions.

In addition, let us assume that there are $K$ input features $X_j(i,t)$, $j=1:K$ that potentially impact the metric of interest $Z$. Some input features, such macroeconomic indicators, can be defined only as time series omitting element specific dependencies. However, these special cases will not limit the problem definition and the applied approach.

Figure 1 illustrates the relational structure of datasets and time series $Z(t)$.

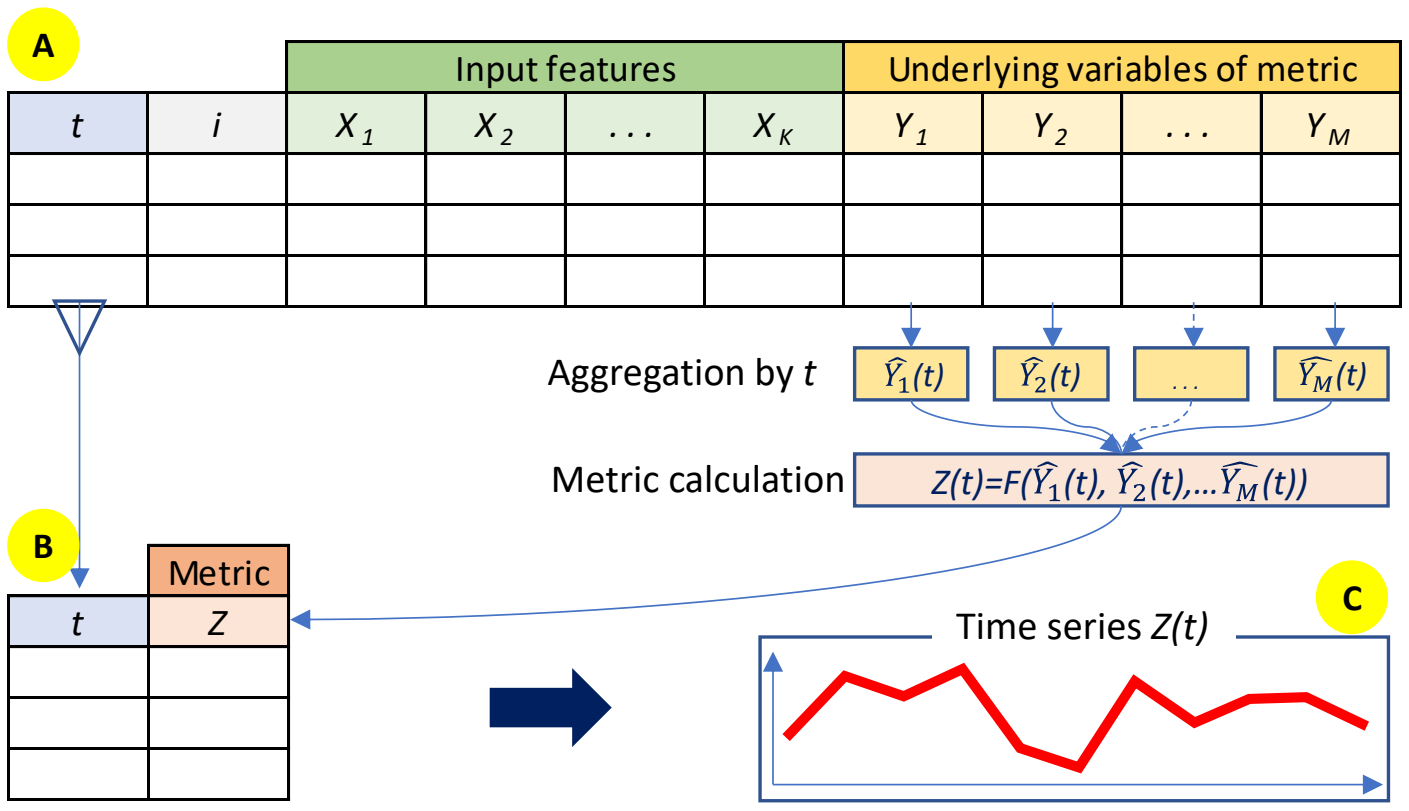

Figure 1. Structure of relational datasets providing input features and underlying variables (A) to calculate the metric of interest (B) and corresponding time series (C)

The problem then is: How can the observed time series of the metric *Z(t)* be interpreted in terms of the contributing factors of variance by knowing input features $X_j(i,t)$ and underlying variables $Y_m(i,t)$ of the metric calculation (1)?

There are several analytical challenges of the defined problem: (1) a non-additive metric, (2) preceding aggregations of underlying variables to calculate the metric, (3) possible nonlinearity of relationships between input features and the target metric, (4) interpretability that eventually requires additive decomposition of the time series, and (5) dimensionality. The latest reflects that the input features and underlying variables used to calculate the metric of interest are formed as two-dimensional arrays along elements *i* and time points *t* while the metric time series is a one-dimensional array.

## 3. Additive Decomposition of Time Series

To address the defined problem and to provide the linear decomposition of *Z(t)* by contributing factors, the following steps have been considered:
(1) segmentations along input features and the underlying variables of the metric of interest followed by aggregations
(2) univariate or joint fittings of the calculated target metric by the aggregated input features on the segmented domains
(3) preliminary screening and transformations of input features by applying the fitted models
(4) aggregations of the transformed features as time series
(5) modelling of the metric time series as combined effects of the aggregated features by applying constrained linear regression

### Step 1: Segmentations of Input Features and Underlying Variables

Two approaches for segmentations along input features and underlying variables are discussed in the paper: univariate and generalized.

Univariate segmentations $(i,t)_j \rightarrow s_j$ are individually driven by each input feature $j=1:K$.

This step assumes robust estimation of metric *Z* that is subject to sufficient sizes of segments aggregating underlying variables $Y_{1:M}$, while not masking the dependencies to the input feature. The latest may happened with segments that are too large.

Univariate segmentations can be performed by different methods, such as percentiles, equally ranged bins of the input features, or unsupervised representations with the defined grid on the latent space.

Generalized segmentation $(i,t) \rightarrow s$ is driven by underlying variables $Y_{1:M}$ that are used as input features in the variational autoencoder. It creates unified segmentation across all input features and underlying variables. The motivation for this approach is driven by the ability to capture metric variability by having underlying variables disentangled on the latent space.

Applications of unsupervised representations using variational autoencoders (β-VAE) (Kingma and Welling, 2014, Higgins *et al*, 2017) for both univariate and generalized approaches are discussed later in the described example.

### Step 2: Univariate or Joint Fittings of the Metric

The next step is univariate or joint fitting, and preliminary screening of *K* input features predicting the target metric *Z* on segmentation levels $s_j$ or *s*, correspondingly:

$$\tilde{Z}(s_j) = G_j(\hat{X}_j(s_j)), j=1:K \quad (2)$$
$$\tilde{Z}(s) = \gamma_0 + \sum_{j=1}^{K} \gamma_j G_j(\hat{X}_j(s)) \quad (3)$$

Modeling (2) can be done by applying traditional regressions, decision trees, or deep neural networks. Fitting the joint model (3), that represents GAM (Hastie and Tibshirani, 1986), can be done by applying Neural Additive Models (NAMs) (Agarwal *et al*, 2021).

Model fitting includes input features screening that rejects some variables having lower than set threshold fitting criteria for traditional approaches, such as $R^2$ or $\chi^2$ (Sarma, 2007), or by applying *L1* regularization for deep neural networks.

Sometimes, metrics are defined within specific ranges, usually [0,1]. To ensure that the modeled values are within the valid ranges, a common approach of a preliminary transformation of the target variable applying functions such as PROBIT or LOGIT followed by the inverse transformation of the modeled values can be used.

Addressing some business-driven requirements, such as monotonicity for continuous variables or ordering for categorical data, constrained optimization can be applied to solve this issue.

In addition, a delayed metric response can be considered at this step. It can be done by forward shifting $X_j(i,t+\Delta t)$, $\Delta t$ = 1, 2, … etc. which maximizes the fitting of the models (2, 3) by period $\Delta t$ as additional parameter. Usually, that simple shift provides good coverage of the dynamic aspects of the metric responses. However, in some cases, it may require fitting of a dynamic response curve $G_j(X_j(\Delta t))$ (Figure 2).

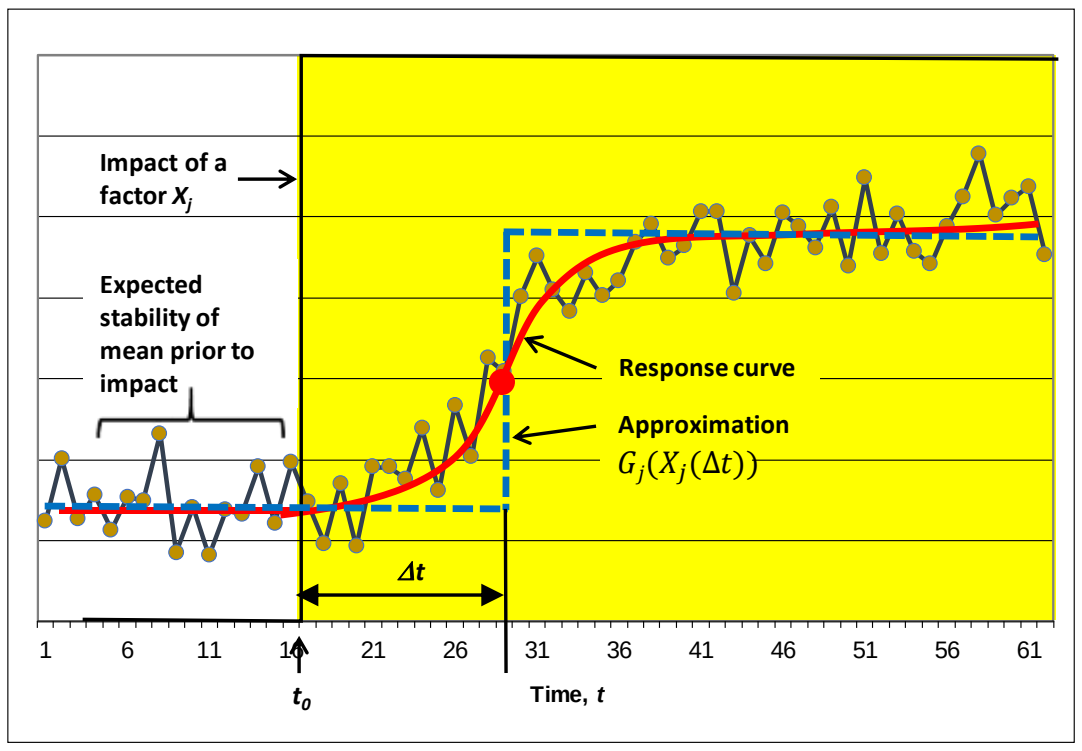

Figure 2. The typical response curve (solid red line) and step function approximation (dashed blue line) on the stepped impact at time $t_0$ (yellow zone)

## Step 3: Transformations of Input Features

After fitting models (2, 3), the population of the modeled values $G_j$ for each element *i* at time *t* is performed:

$$X_j(i,t) \xrightarrow{G_j} G_j(X_j(i,t)) \quad (4)$$

This transformation is performed only for input features that survived screening in step 2. Diagram of the univariate transformations of input features is presented in Figure 3.

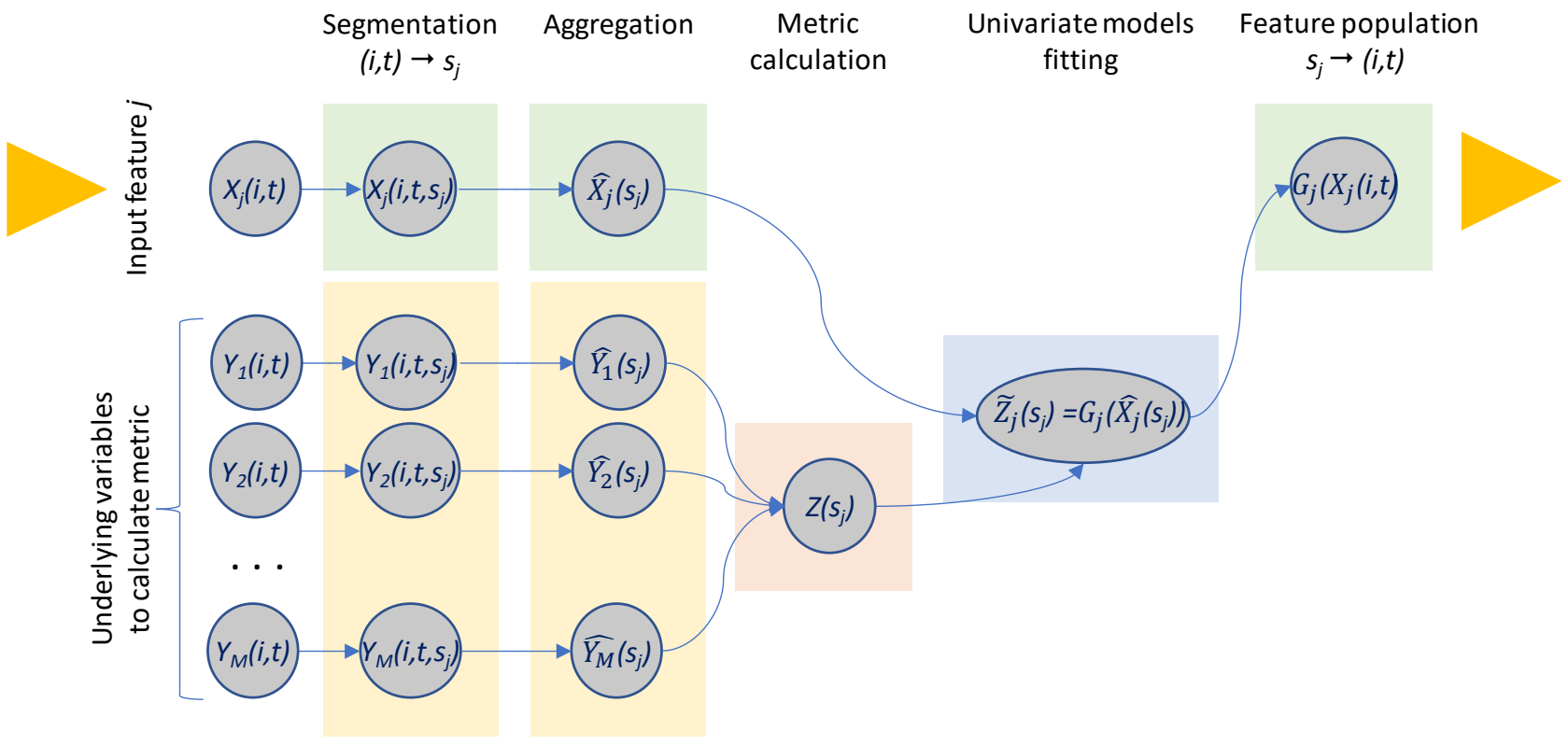

Figure 3. Univariate transformations of input features (*j=1:K*)

Illustration of the generalized transformation is shown in Figure 4. It should be noted that this approach allows for NAMs application (Agarwal *et al*, 2021) of the joint model fitting across all input features.

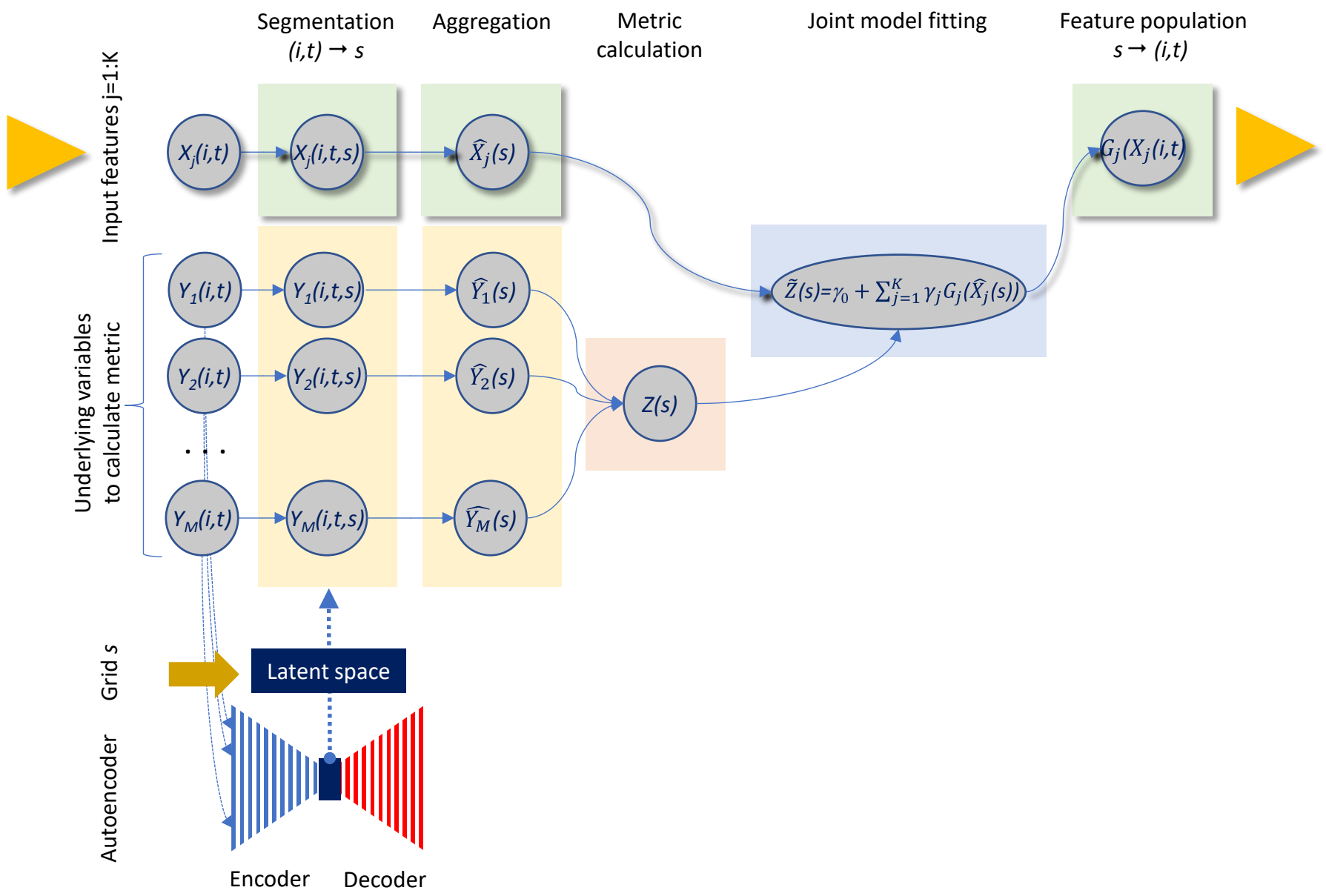

Figure 4. Joint transformations of input features

The relationship to target transformation is a widely used approach, for example, when dealing with binary target variables (Sarma, 2005; Sarma, 2007) or applying smooth functions in the Generalized Additive Models (GAM) (Hastie and Tibshirani, 1986). Transformations to binary target variables usually include calculations of Weight of Evidence (WOE) followed by logistic regression modeling (Ondrej, 2008; Siddiqi, 2012). Combining univariate decision trees with logistic regression incorporates non-linear relationships between input features and the target variable in predictive modeling (Sarma, 2005). However, addressing the time series of non-additive metrics, additional steps are required: first segmentations followed by aggregations producing time series for transformed and screened input features to be used in the final modelling step.

## Step 4: Aggregations of Transformed Features as Time Series

The next step is to aggregate transformed features as time series. The motivation for that step is directed by the dimensional challenge of having two-dimensional arrays of input features and a one-dimensional array of the metric time series.

Depending on the metric form, the aggregation step includes either simple averaging:

$$\hat{G}_j(t) = \frac{1}{N(t)} \sum_{i \in P(t)} G_j(X_j(i,t)),\, j = 1{:}K \qquad (5)$$

or weighted averaging. For example, considering the loss-to-balance ratio, the aggregation is weighted by the balances:

$$\hat{G}_j(t) = \sum_{i \in P(t)} B(i,t) G_j(X_j(i,t)) \,/\, \sum_{i \in P(t)} B(i,t) \qquad (6)$$

where *B(i,t)* is an outstanding balance of *i* account at time *t*.

In addition, the normalization of $\hat{G}_j(t)$ can be applied towards factor comparability across the monitoring period.

## Step 5: Time Series Modeling

The last step of the approach to fit the model of the metric time series. Considering the interpretability of the model, i.e., the ability to estimate effects of the contributing factors, time series modeling is limited to a constrained linear regression of the transformed and aggregated time series $\hat{G}_j(t)$:

$$\tilde{Z}(t) = \beta_0 + \sum_{j=1}^{K} \beta_j \hat{G}_j(t) \qquad (7)$$

having positive parameters $\beta_j \geq 0$.

The final model (7) provides additive decomposition of the metric time series and supports interpretability concerning contributions of the input features. It can be seen as a modified GAM model (Hastie and Tibshirani, 1986) applied for time series decomposition.

The fitting of the final model can be set either as a constrained optimization forcing parameters $\beta_j$ to be non-negative with backward, forward, or stepwise feature screening, or by applying a deep neural network with Lagrangian regularization against negative parameters.

The constraint, forcing positivity of parameters $\beta_j \geq 0$, confines overfitting that could potentially be caused by collinear $\hat{G}_j(t)$ input features. The rationale behind that constraint is that the target metric has been used to transform the original input feature during step 3 of the approach, and, therefore, the time series $\hat{G}_j(t), j = 1{:}K$ assume positive correlations to the target.

Individual contributions *H(i,t)* of elements *i* of the entity can be learned by applying the fitted model (7) to the transformed input features $G_j(i,t)$:

$$H(i,t) = \beta_0 + \sum_{j=1}^{K} \beta_j G_j(X_j(i,t)) \quad (8)$$

The diagram illustrating aggregation of the transformed input features as multiple time series and final model fitting (7) is shown in Figure 5. It should be noted that some transformed features may be omitted due to screening during fittings (2, 3). Furthermore, significant impact on the target metric observed in fittings (2, 3) is a necessary but not a sufficient condition for a factor to survive in the final model (7).

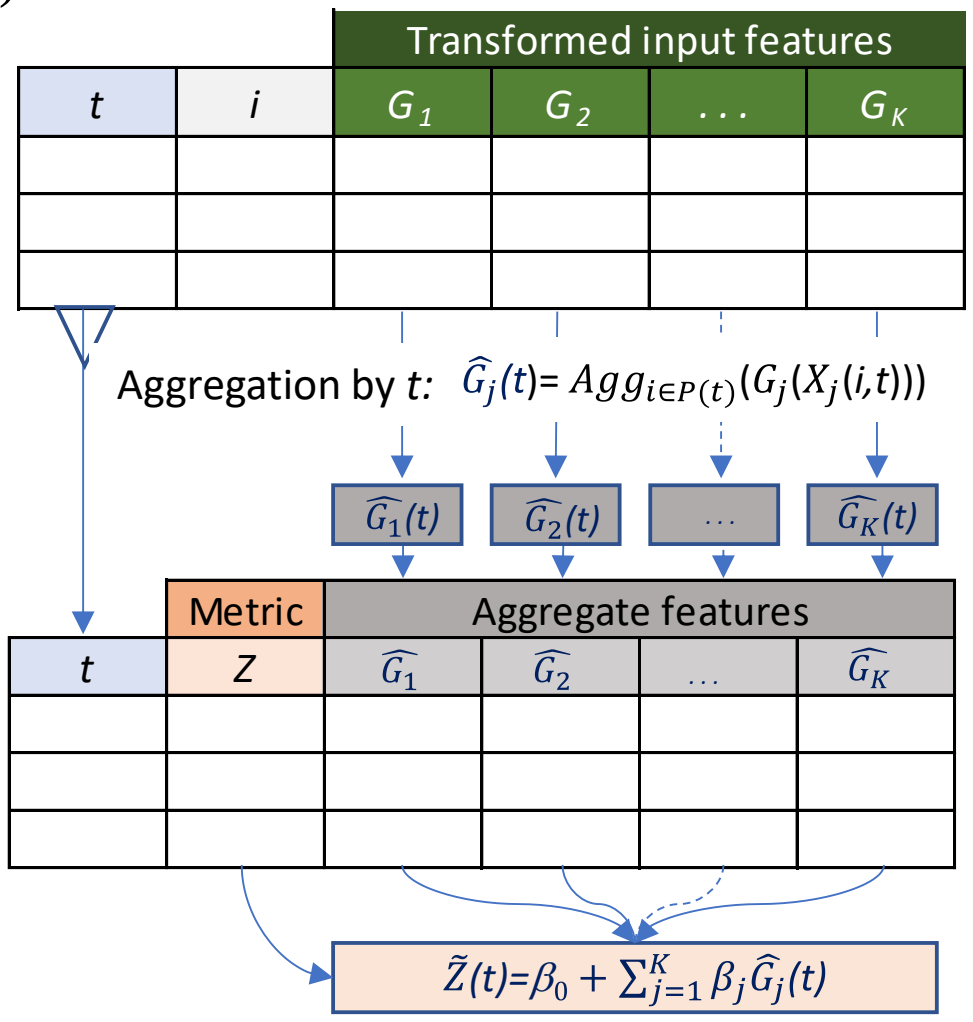

Figure 5. Diagram illustrating aggregation of the transformed input features as multiple time series and final model fitting

## 4. Metric Linearization by Numerical Differentiation

Numerical differentiation can be applied to linearize the non-additive metrics and provides an alternative path for the first two steps of the approach. Numerical differentiation emphasises local variances similar to the LIME technique (Ribeiro, 2016).

As an example, let us consider the most common non-additive metric that equals a simple ratio: $Z(t) = \hat{Y}_1(t)/\hat{Y}_2(t)$.

Applying differentiation along time *t*, the metric period-to-period changes can be presented as:

$$\Delta_t Z(t) = \frac{1}{\hat{Y}_2(t)} \Delta_t \hat{Y}_1(t) - \frac{\hat{Y}_1(t)}{\hat{Y}_2^2(t)} \Delta_t \hat{Y}_2(t) \quad (9)$$

where operator $\Delta_t$ represents numerical differentiation along time *t*, for instance: $\Delta_t Z(t) = Z(t) - Z(t-1)$

Linearization of (9) can be done by the following approximation:

$$\Delta_t Z(t) \cong C_1 \cdot \Delta_t \hat{Y}_1(t) - C_2 \cdot \Delta_t \hat{Y}_2(t) \quad (10)$$

where $C_{1,2}$ are constants that can be either calculated based on mean values of the corresponding terms ($C_1 \cong T / \sum_{t=1}^{T} \hat{Y}_2(t)$; $C_2 \cong T \cdot \sum_{t=1}^{T} \hat{Y}_1(t) / \left(\sum_{t=1}^{T} \hat{Y}_2(t)\right)^2$) or fitted parameters.

The approximation (10) assumes that the second-order deviations can be ignored.

Equation (9) allows for approximation of the metric time series *Z(t)* by summarization along time *t* as a reverse process to differentiation:

$$Z(t) \cong Z(0) + \sum_{\tau=1}^{t} (C_1 \cdot \Delta_t \hat{Y}_1(\tau) - C_2 \cdot \Delta_t \hat{Y}_2(\tau)),\ t \geq 1 \quad (11)$$

or, after changing order of summations:

$$Z(t) \cong Z(0) + \sum_{i \in P(t)} \sum_{\tau=1}^{t} (C_1 \cdot \Delta_t Y_1(i,\tau) - C_2 \cdot \Delta_t Y_2(i,\tau))$$

and finally:

$$Z(t) \cong Z(0)^* + \sum_{i \in P(t)} \big(C_1 \cdot Y_1(i,t) - C_2 \cdot Y_2(i,t)\big) = Z(0)^* + \sum_{i \in P(t)} L(i,t) \quad (12)$$

where $L(i,t) = C_1 \cdot Y_1(i,t) - C_2 \cdot Y_2(i,t)$ in the approximation (12) is a linear form on element level $i$. This property allows for direct application of univariate:

$$\tilde{L}(i,t) = G_j(X_j(i,t)),\ j=1{:}K \quad (13)$$

or joint GAMs or NAMs models:

$$\widetilde{H}(i,t) = \beta_0 + \sum_{j=1}^{K} \beta_j G_j(X_j(i,t)) \quad (14)$$

both on an element level with the ability to aggregate results for an entity level.

The concerning issue of this approach is accuracy of summation along time $t$ in the reverse process (11) that accumulates errors of the approximation.

## 5. Synthetic Example

To illustrate the approach, let us consider a hypothetical retail credit portfolio with a metric of interest that equals to a loss-to-balance ratio (simply, loss rate):

$$Z(t) = \frac{\sum_{i \in P(t)} Y_1(i,t)}{\sum_{i \in P(t)} Y_2(i,t)}$$

where $Y_1$ is the loss that for simplicity equals the outstanding balance at default of an account $i$ or zero otherwise, and $Y_2$ is the outstanding account balance.

Let us assume that the portfolio has the following properties:

- It is a fast-growing portfolio with no attrition
- There are three distinct customer segments 'A', 'B', and 'C', which can be identified at the time of acquisition. These segments have significant and long-term discriminatory power concerning the risk-return qualities of the acquired accounts
- Outstanding account balance ($Y_2$) is a stochastic function of account characteristic categorized by three customer segments, tenure of account, and seasonality
- Account probability of default is a stochastic function of the unemployment rate, account characteristic categorized by three customer segments, and tenure of account
- During the monitoring period, there were two changes on the portfolio, both concerning the acquisition:
  (I) managerial decision to decrease acquisition of the segment 'C' after time $t_1$.
  (II) regulatory requirement to decrease acquisition of segments 'B' and 'C' after time $t_2$.
  The first change has been driven by observing underperforming results for an aggressively growing segment "C". The second change was a reaction of a regulator due to an economic downturn preventing future credit quality degradation. It impacted the acquisition of segments "B" and "C". As a result, better acquisition quality should be expected with a fewer number of acquired accounts
- The following are assumed input features of the portfolio loss rate variance: macroeconomic environment described by the national unemployment rate ($X_1$), acquisition characteristic identified by three segments 'A', 'B', and 'C' ($X_2$), tenure of account ($X_3$), acquisition change caused by managerial decision ($X_4$), acquisition change triggered by the regulatory requirement ($X_5$), and seasonality ($X_6$).

The above specified properties of the portfolio, environment, and input features are for simplicity of the example and to illustrate the described approach.

Simulated monthly loss rates are shown in Figure 6 (solid blue dots). This time series represents the portfolio metric of interest $Z(t)$.

The first four steps of the approach are: segmentations, fittings to the target metric, transformations, and aggregations for each factor $X_{1\text{-}6}$.

The transformation of the simulated monthly unemployment rates $X_1$ to the target metric $Z$ includes fitting a non-linear model with a monotonicity constraint as shown in Figure 7, a. The monotonicity constraint ensures the following relationship: the higher the unemployment rate, the higher the loss rate of the retail portfolio.

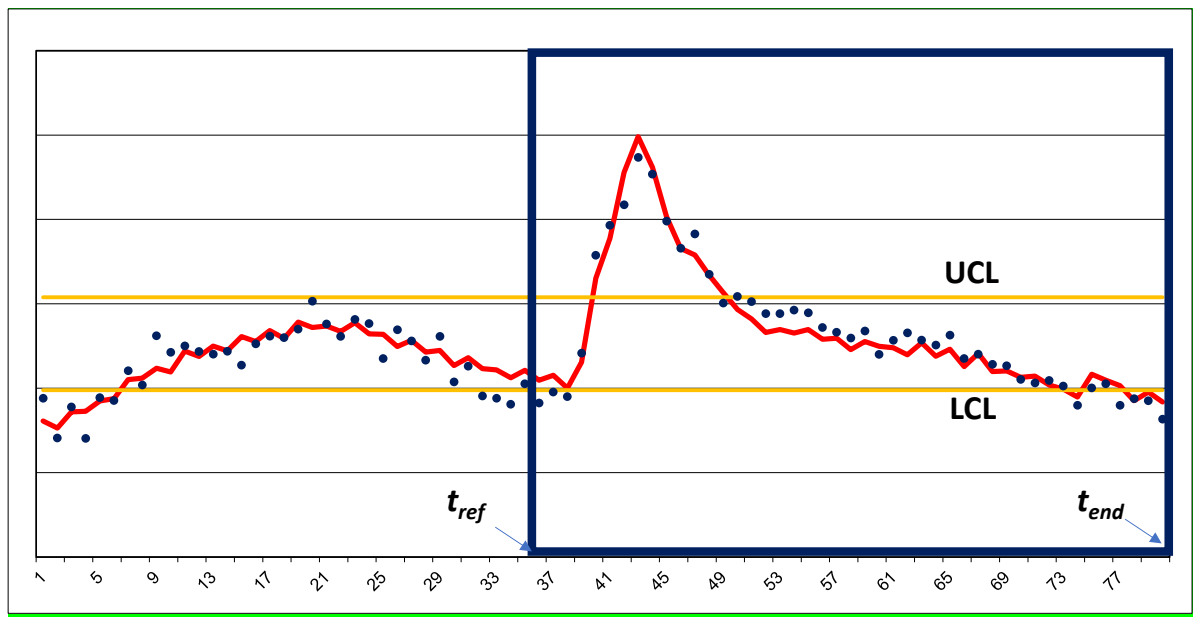

Figure 6. Simulated portfolio monthly loss rates (solid blue dots).

As mentioned above, the lagged effects can be captured by shifting $X_j(i,t+\Delta t)$, $\Delta t = 1, 2, \ldots$ etc. periods forward. However, for simplicity, this possibility is not addressed in the example.

Another example of the univariate dependence between the tenure of account ($X_3$) and the target metric is shown in Figure 7, b. This typical chart blends different acquisition periods that reduces noise from other input features. Furthermore, when considering acquisition as a step-shaped impact, the curve in Figure 7, b is a special case of a dynamic response curve $G(X_j(\Delta t))$ that has been discussed earlier (Figure 2), where $\Delta t$ is the tenure of account ($\Delta t \geq 0$).

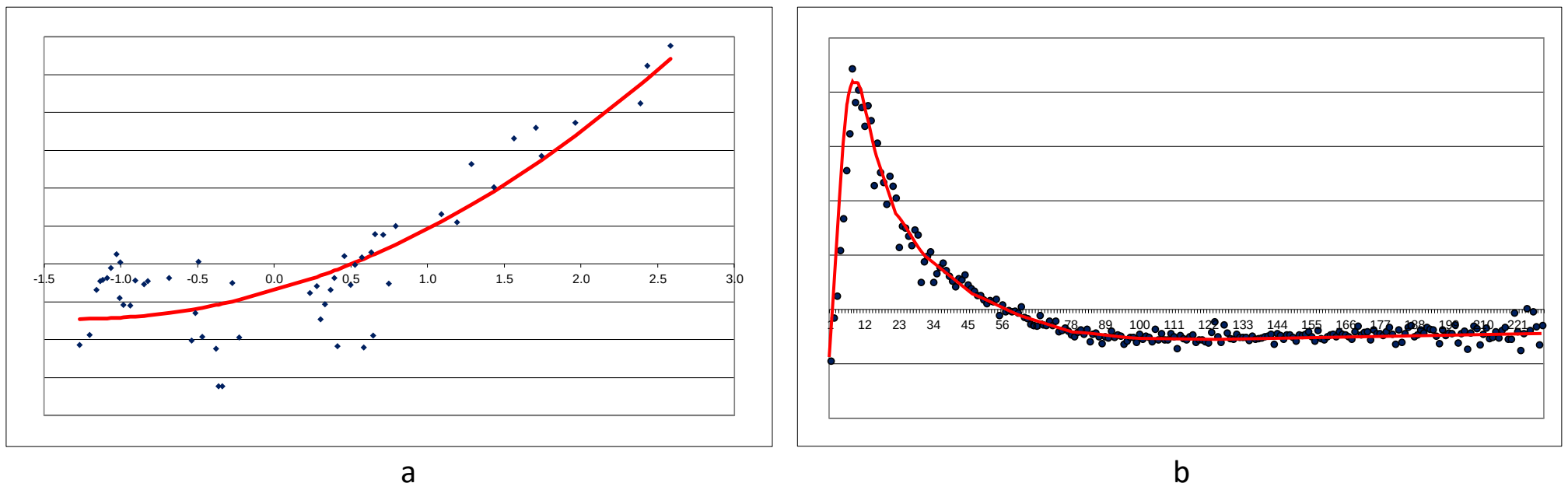

Figure 7. Examples of the fitted univariate non-linear models (solid red lines) of the loss rate against (a) the unemployment rate $X_1$ with monotonicity constraint, and (b) the tenure of account $X_3$

## 5.1. Application of Unsupervised Variational Autoencoders

Application of unsupervised variational autoencoders β-VAE can be used to drive the segmentation step. Furthermore, it can support both univariate and generalized segmentations. The latest means unified segmentation across all input features $X_{j=1:K}$ and underlying variables $Y_{j=1:M}$. and allows for implementation of the NAMs architecture (Agarwal *et al*, 2021) to jointly train $G_{j=1:K}$ models (3).

The rationale of applying autoencoders is driven by the following diagram:

**Latent Space Representation ➔ Regular Grid Application ➔ Segmentation ➔ Aggregation**

The latent space representation by β-VAE supports disentanglements of the input features on the latent space and it is forced towards isotropic normal distribution (Kingma and Welling, 2014, Higgins *et al*, 2017).

The normally distributed latent space is then transformed into a uniformed one by applying a normal cumulative distribution function for the latent space axes: (-∞; +∞) → (0; 1). The uniformed latent space allows for application of regular grids, such as square, and essentially segmentation.

Example of the univariate segmentation based on the latent space representation for tenure of account $X_3$ is shown in Figure 8.

Time $t$ has been added as an additional input feature of the β-VAE. It provides a possibility to control stability of the input feature $X_3$ distribution over time $t$ on the latent space and, consequently, time dependence of the model $G_3$.

It can be observed that the input feature of tenure of account $X_3$ has good disentanglement along the ordinate axis of the latent space while gradient of time $t$ aligns with abscissa (Figure 8). It means orthogonal disentanglement between input feature of tenure of account $X_3$ and time $t$.

In the shown example, the regular square 10x10 grid on the latent space has been applied as segmentation $s_3$ with the following aggregation (mean) of the input feature $\hat{X}_3$ and metric calculation. It provides necessary data to train the univariate model $\tilde{Z}(s_3) = G_3(\hat{X}_3(s_3))$.

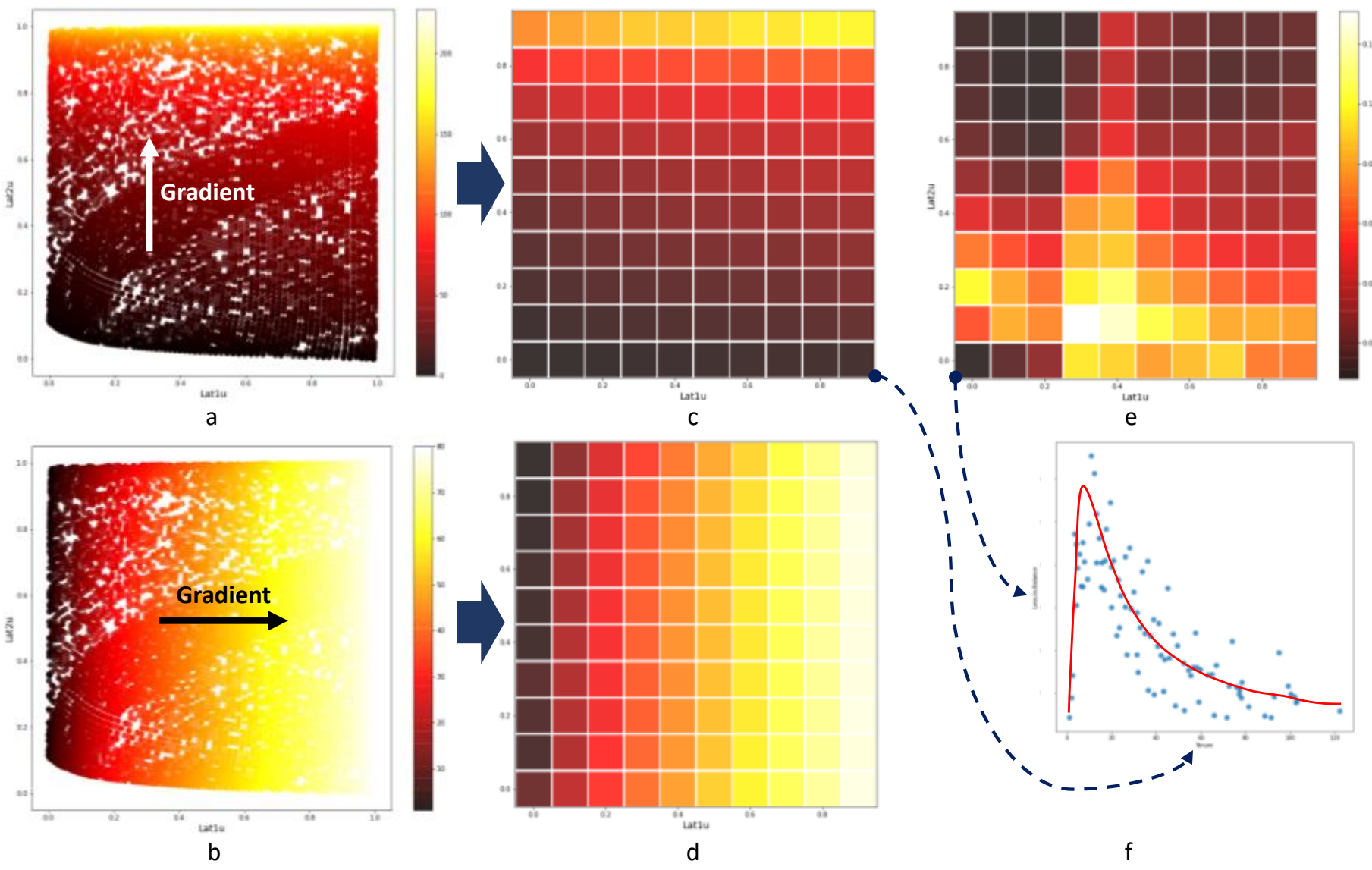

Figure 8. Unsupervised representations on the 2D latent space by β-VAE:

Mapping of the input feature $X_3$ of tenure of account (a) and time $t$ (b); and their segmented mean estimations based on the defined 10x10 square grid (c) and (d), respectively; (e) calculated loss-to-balance metric based on the loss ($Y_1$) and balance ($Y_2$) aggregations across the defined grid; (f) loss-to-balance ratio versus tenure of account based on the defined segmentation

Example of a generalized representation of underlying variables $Y_1$ (loss) and $Y_2$ (balance) as β-VAE input features is shown in Figure 9.

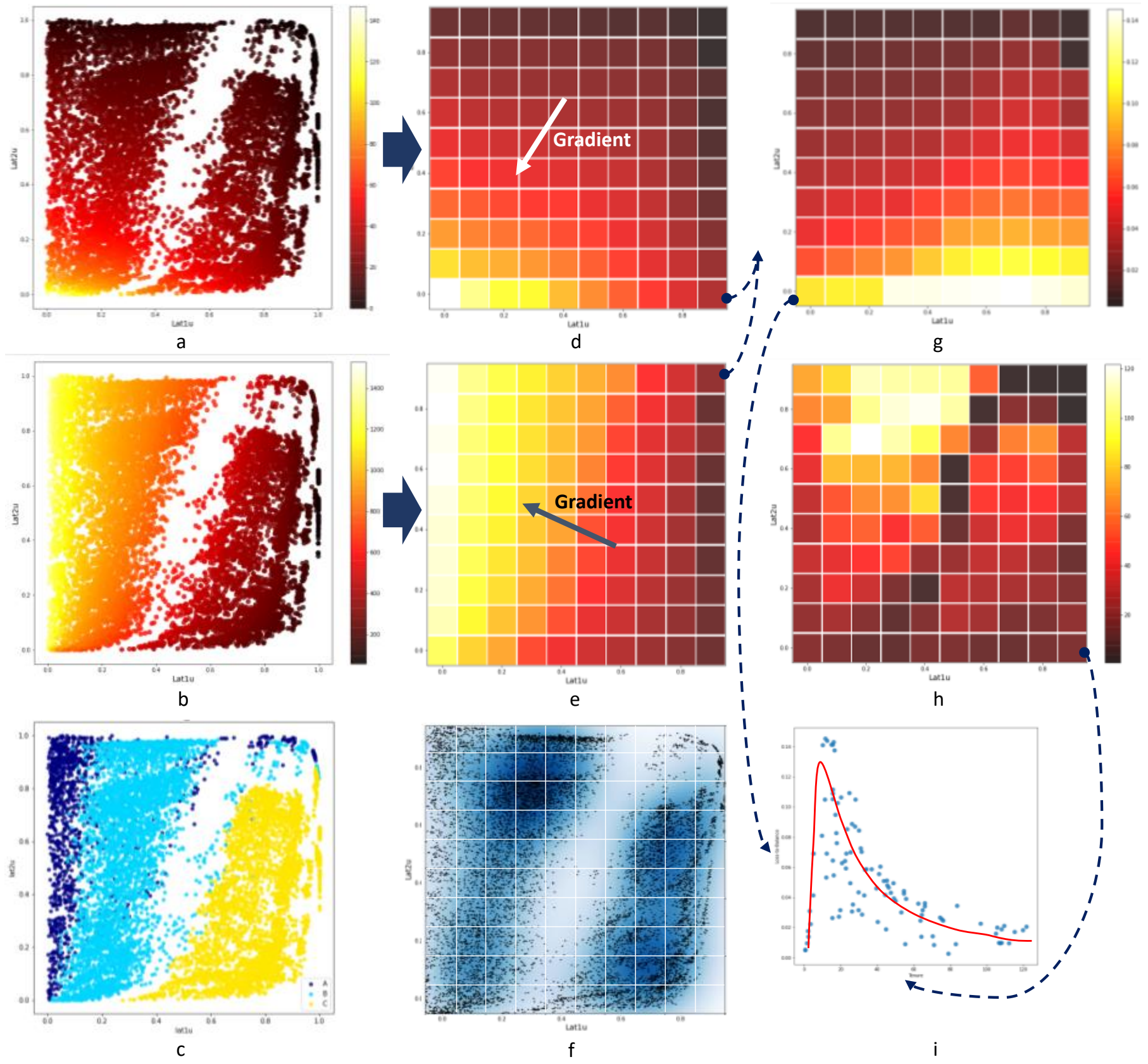

Figure 9. Unsupervised generalized representations on the 2D latent space by β-VAE:

Mapping of underlying variables: loss $Y_1$ (a) and balance $Y_2$ (b), and their segmented mean estimations (d) and (e) based on the defined square 10x10 grid, respectively; (c) mapping of input feature acquisition characteristic $X_2$; (g) calculated loss-to-balance metric based on the defined grid; (h) estimated (mean) tenure $X_3$ input feature based on the defined grid; (f) density distribution of the representation on the latent space; (i) loss-to-balance ratio versus tenure of account based on the defined segmentation

It can be observed that there are strong disentanglements of the underlying variables $Y_1$ and $Y_2$ on the latent space (Figure 9, a and b) and, consequently, their aggregated values (sums) $\hat{Y}_1(s)$ and $\hat{Y}_2(s)$ on the applied regular square 10x10 grid (Figure 9, d and e). The calculated metric $Z(s) = \hat{Y}_1(s)/\hat{Y}_2(s)$ has a distinct pattern on the defined grid $s$ (Figure 9, g).

As an example, the mapping of the aggregated (mean) input feature of tenure of account $\hat{X}_3(s)$ is presented in Figure 9, h. Similarly, other input features can be mapped on the defined grid. For instance, mapping of the acquisition characteristic $X_2$ is shown in Figure 9, c.

Encoding based on underlying variables supports generalized segmentation across the latent space and allows for joint training of the model (3).

It can be noted that density distribution is not perfectly unformed (Figure 9, f). To ensure robustness of the aggregations, it may require density-based filtering, keeping only cells of the defined grid with sample sizes above the specified threshold level. In addition, tuning of hyperparameter $\beta$ of β-VAE can be applied to control distribution on the latent space against the accuracy of the encoding-decoding.

## 5.2. Time Series of Transformed Input Features

After populating the transformed data of input features $G_j(i,t)$, the aggregated and normalized time series $\hat{G}_j(t)$ have been calculated and shown in Figure 10. The blue areas (below zero) represent periods, when the factors have positive (decrease) impacts on the loss rate, and the red areas (above zero) represent downturn periods with negative (increase) impacts.

It should be noted that the identified periods with positive and negative impacts on the portfolio loss rate by input features are valid for the specified monitoring period $T$. The paper later discusses the possibility to study impacts within the different sub-monitoring periods providing a linear model (7).

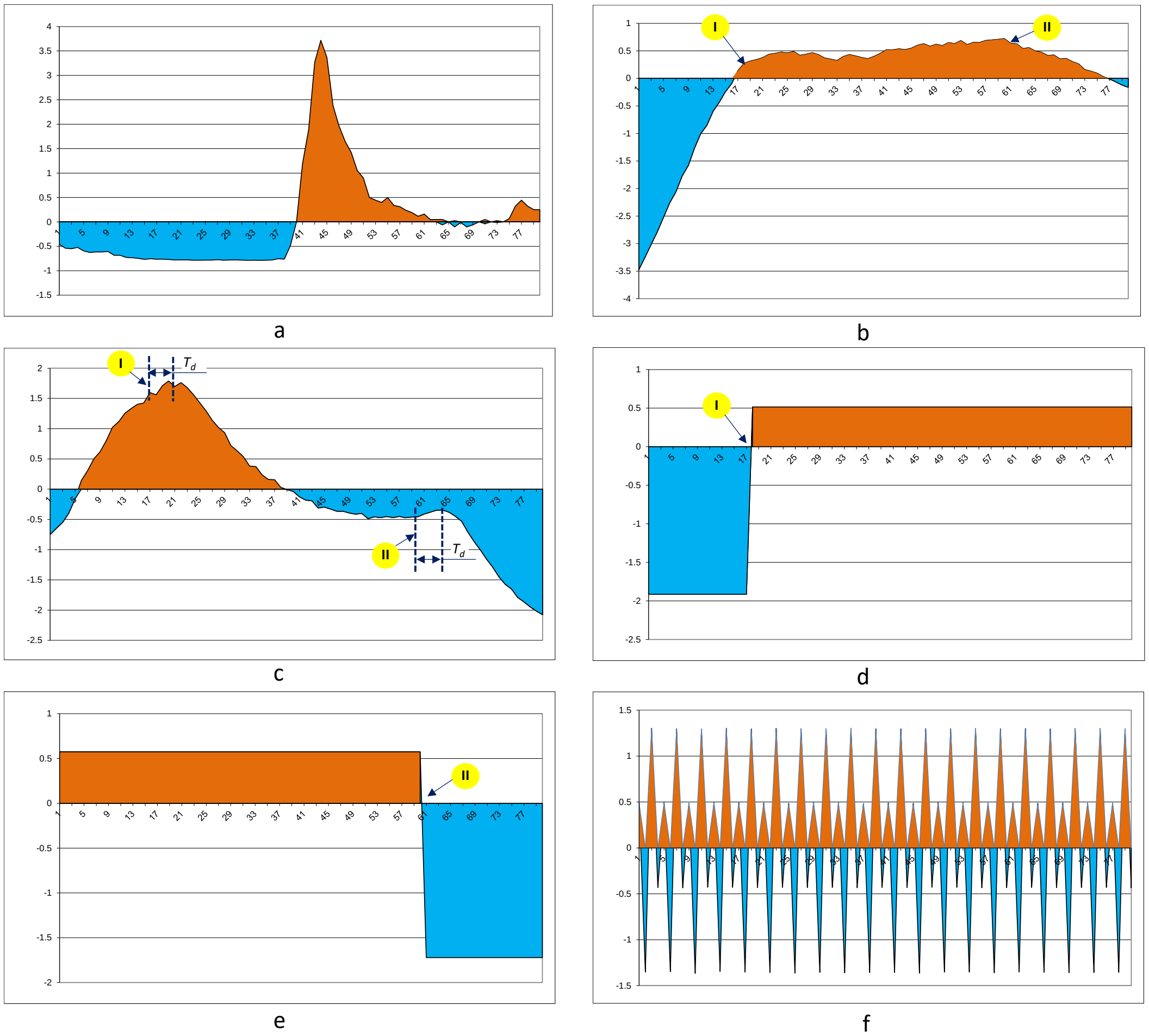

Figure 10. The aggregated and normalized time series $\hat{G}_{1:6}(t)$ of the transformed input features:
(a) unemployment rate $X_1$; (b) acquisition characteristic $X_2$; (c) tenure of account $X_3$; (d) managerial decision factor ($X_4$) that occurred after time $t_1$ (point I); (e) regulatory requirement factor ($X_5$) that occurred after time $t_2$ (point II); (f) seasonality factor $X_6$

As stated above, accounts are grouped into three acquisition characteristic segments 'A', 'B', and 'C'. Practically, it requires preliminary segmentation modeling using account level input features, such as an account's risk level, product attributes, loan-to-value ratios for secured credits, etc. - all known at the time of acquisition. This type of modeling is a typical acquisition score fitting, for example, credit risk scorecard (Siddiqi, 2012). Reducing noise from other input features, the model target metric should reflect the mid-term cumulative outcome at a specified tenure of account. In addition, the training dataset should include accounts with acquisition periods spread out (ideally randomly) throughout the macroeconomic cycle. The trained model can then be used to score each account reflecting acquisition characteristic.

In our example, acquisition characteristic is a static account attribute that is categorized by the three defined segments. However, the number of accounts per each customer segment vary over time.

The negative impact of the acquisition characteristic after the first quarter of the monitoring period (red area) can be explained by an increased proportion of the customer segment "C" in the portfolio - the fastest growing underperforming segment (Figure 10, b). However, at the end of the monitoring period, the acquisition characteristic has reached the neutral level due to managerial (point 'I') and regulatory (point 'II') changes.

It can be noted that an aggressive acquisition strategy has caused an elevated loss rate (red area zone in Figure 10, c), even though it was the managerial decision to cool down the acquisition after time $t_1$ (point 'I'). This example shows that the effect of the tenure of account has significant inertia since this factor cannot be changed immediately and represents cumulative effects of the acquisition strategy.

Another significant change of the $\hat{G}_3(t)$ curve has been caused by the implementation of the regulatory requirement (input feature $X_5$) after time $t_2$, which had an additional beneficial impact on the portfolio loss rate after point 'II' in Figure 10, c.

In the example, transformations of input features $X_4$ (managerial decision) and $X_5$ (regulatory requirement) are based on simple calculations of metric $Z$ for periods before and after the implementation dates.

It can be learned that the managerial decision to slow down acquisition of segment "C" leads to the elevated loss rate (red area) in Figure 10, d. However, as it will be discussed later, this conclusion is ambiguous considering a very high correlation between this factor and the acquisition characteristic.

As expected, after implementing the new regulatory requirement at point II, the portfolio loss rate has been improved (blue area in Figure 10, e).

In the synthetic example, both time series $\hat{G}_4(t)$ and $\hat{G}_5(t)$ are step functions (Figure 10, d and e). However, in the real world, these time series may have different shapes providing that regulatory or managerial effects can be account specific and be spread out over time. Also, the transformation step may incorporate dynamic aspects of the metric responses as mentioned earlier (Figure 2).

Seasonality factor $X_6$ has been transformed into $G_6$ as simple monthly averages of the metric $Z$. It is no surprise that the aggregated time series $\hat{G}_6(t)$ has a periodic pattern that is presented in Figure 10, f.

Finally, providing time series $\hat{G}_{1-6}(t)$ of all the input features (Figure 10), it is possible to fit a linear regression model (7) constrained to positive parameters $\beta_{1-6}$. The final model after backward elimination includes three input features: unemployment rate $\hat{G}_1(X_1)$, acquisition characteristic $\hat{G}_2(X_2)$, and tenure of account $\hat{G}_3(X_3)$. The time series of the fitted model (7) is presented in Figure 6 as a red solid line ($R^2_{ADJ}$=92%).

The major concern of having potentially $K$ predictive time series $\hat{G}_j(t)$, while the monitoring period is limited to $T$ months, is the available degree of freedom. It possesses the risk of model overfitting.

In the example, there is a strong positive correlation between the acquisition characteristic and the managerial decision (Figure 10, b and d). It reflects the fact that the reason of the managerial decision was to improve acquisition quality. It can be observed that although the acquisition characteristic of the aggregated time series has a negative impact on the loss rate after the first quarter of the monitoring period, it has a significant trend of slowing down the negative impact after point 'I' (Figure 10, b). This trend can be attributed to the managerial decision of decreasing segment "C" acquisition.

Also, there is moderate positive correlation between the regulatory change and the tenure of account (Figure 10, c and e). That can be explained by the impact of the regulatory requirement on the acquisition volume and eventually on the distribution of accounts across life stages.

It should be noted that the numerical differentiation approach (13) provides quite similar results for all survived input features' $\hat{G}_{1-3}(t)$ curves shown in Figure 10, a, b, and c with $R^2$>91%.

## 5.3. Interpretation of Input Features Effects

The additive model (7) allows for studies of normalized input features' contributions $\beta_j \hat{G}_j(t)$ to the monthly loss rate changes around the average level $\beta_0$ for the monitoring period $T$ noting that $\sum_{t=1}^{T} \beta_j \hat{G}_j(t) = 0$.

The chart in Figure 6 suggests that the trained model (red curve) has quite a good fit ($R^2_{ADJ}$=92%) of actual monthly variations of the loss rate (blue points). Considering the three survived input features of the final model (7) and their normalized parameters $\beta_j$ along with the aggregated time series $\hat{G}_j(t)$ charts, it is possible to analyze the effect of input features on the portfolio loss rate:

- The loss rate has the highest sensitivity to unemployment rate ($X_1$) with a positive (decrease) impact during the first half of the monitoring period and a negative (increase) or neutral impact after (Figure 10, a). These effects are synchronized with the macroeconomic cycle.
- The next important factor is the tenure of account ($X_3$), which has a very strong increasing trend during the first quarter of the monitoring period and then a decreasing trend after (Figure 10, c). It was hitting the red zone due to a very aggressive acquisition strategy and was coming back to having a positive impact (blue zone) during the second half of the monitoring period. The curve has two distinct trends caused by the managerial decision (factor $X_4$) and implementation of the regulatory requirement (factor $X_5$). These changes can be clearly observed in Figure 10, c, where points 'I' and 'II' represent implementation periods of the managerial and regulatory changes, respectively. Some delays ($T_d$) between implementations and effects of about five months are due to the almost neutral level of the loss rates for new accounts with tenure under five months (Figure 7, b).
- The acquisition characteristic ($X_2$) has low sensitivity. As discussed above, there were two assignable changes of the aggregated time series (points 'I' and 'II' in Figure 10, b). The first was caused by the managerial decision of decreasing the acquisition of the underperforming segment 'C' and the second reflects decreased volume of segments 'B' and 'C' that was triggered by the regulatory requirement.

In contrast to the Age-Period-Cohort (Yang and Land, 2013) or dual-time dynamics methodology (Breeden, 2007; Breeden *et al*, 2008), which granularly model vintage (cohort) performances by functions of age, calendar time, and quality of vintage, the five-step approach decomposes time series of the metric as a linear form of time series of survived input features on an entity (portfolio) level. Thus, for developed portfolios having stable distributions of accounts along life stages, the tenure input feature will normally be rejected in the final model (7) due to low variance of the corresponding time series $\hat{G}_3(t)$. It will happen even though this factor may be a powerful predictor considering vintage performance (Figure 7, b). In this case, the trained model (7) cannot be used for predictions assuming severe changes in acquisition volume strategies. It should be noted as a limitation of the entity level decomposition approach considering forward-looking applications.

## 6. Study of Performance Against the Reference Period

The discussed approach allows for performance studies to be done against the reference periods. The simplicity of studies against the defined reference point $t_{ref}$ is due to an additive form of the model (7), which allows for a simple transformation:

$$\tilde{Z}(t) = \tilde{Z}(t_{ref}) + \sum_{j=1}^{K} \beta_j (\hat{G}_j(t) - \hat{G}_j(t_{ref})) \quad (15)$$

where both $t$ and $t_{ref}$ are within the monitoring period that has been used to fit the model (7).

The estimated contribution effect of input feature $j$ on performance during the period between $t_{ref}$ and $t$ equals:

$$Eff_j(t, t_{ref}) = \beta_j (\hat{G}_j(t) - \hat{G}_j(t_{ref})), t > t_{ref} \quad (16)$$

Let us consider the monitoring period of interest [$t_{ref}$; $t_{end}$] that is shown in Figure 6 as a blue rectangle. Figure 11 illustrates the contributions of three input features on the loss rate changes that occurred during that monitoring period.

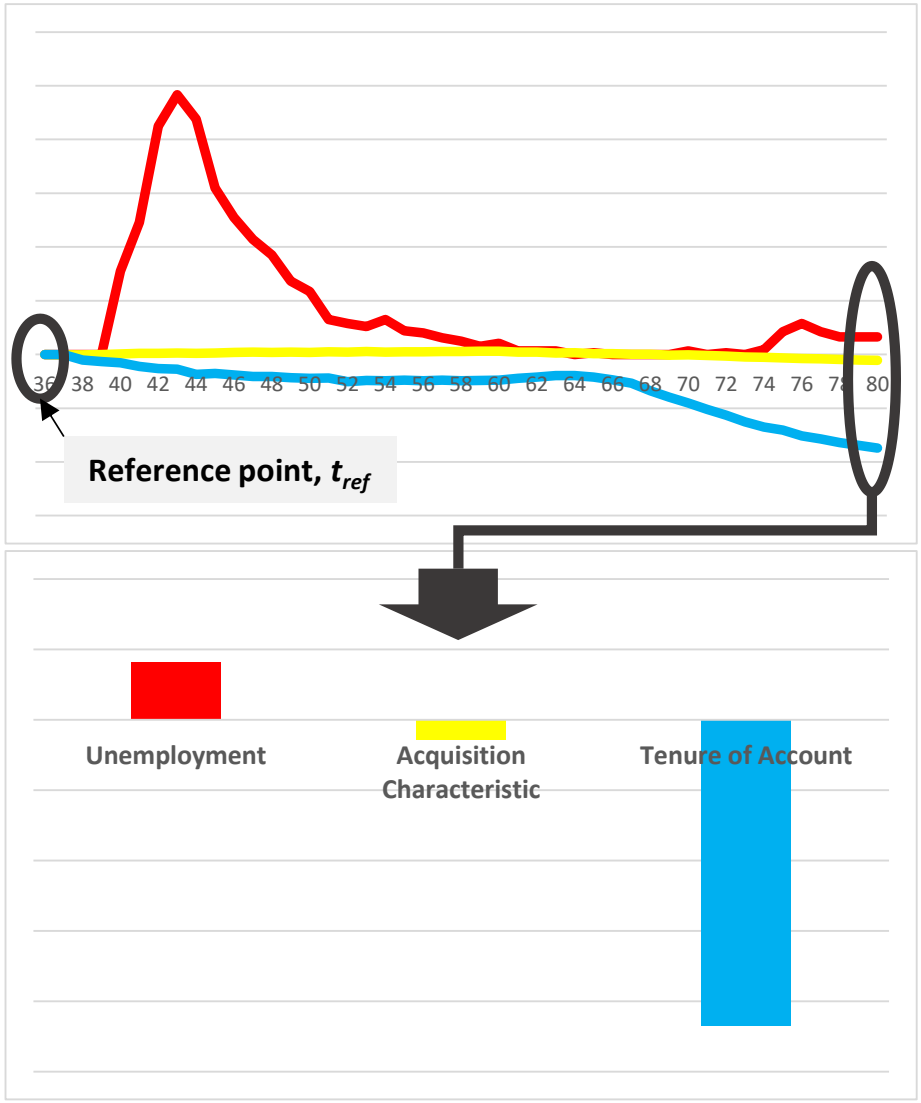

Figure 11. Effects of the survived input features at the end of the monitoring period $t_{end}$ compared to the reference period $t_{ref}$

The portfolio loss rate as of $t_{end}$ is similar to the rate at the reference point $t_{ref}$ (Figure 6). However, two changes have occurred during [$t_{ref}$, $t_{end}$] period:

(1) significant increase in portfolio maturity. It causes a decreasing effect on the loss rate (blue line and bar in Figure 11)

(2) macroeconomic downturn and as a result, elevated unemployment. It causes an increasing effect on the loss rate (red line and bar in Figure 11).

The change in the mix of the acquisition characteristic segments in the portfolio during the monitoring period [$t_{ref}$; $t_{end}$] has an insignificant impact on the loss rate (yellow line and bar in Figure 11).

Since the addressed approach can be applied to distinct portfolio segments, it is possible to map input feature effects (16) on a segment level with sufficient sample sizes. It will provide a better granularity on a factor's effects identifying sensitive or robust segments.

## 7. Statistical Process Control for Residual Time Series

The Statistical Process Control (SPC) technique provides signals in case the stochastic process is out-of-control. Out-of-control means that there are special causes of process variation, while in-control assumes only the presence of common sources of variation (Montgomery, 2007). There are different approaches detecting out-of-control states, providing that the time series in these states is subject to significant trends, shifts, or cycles of either means or variances.

Simple Individual and Moving Range (IX & MR) SPC charts for residuals time series $R(t) = Z(t) - \tilde{Z}(t)$ allow for control of the portfolio metric. If the $R(t)$ time series is in-control, then it means that there are no special causes present, other than the survived input features of the fitted model $\tilde{Z}(t)$ (7).

The IX SPC includes testing that all values of $R(t)$ are within the lower (LCL) and upper (UCL) control limits (Woodall and Faltin, 1993; Montgomery, 2007):

$$LCL = \bar{R} - 2.66 * \overline{MR}$$
$$UCL = \bar{R} + 2.66 * \overline{MR}$$

where $\bar{R}$ is the average of the residual time series $R(t)$ over the monitoring period $T$, in our case $\bar{R} = 0$, and $\overline{MR}$ is the average of the time series moving ranges. It should be noted that there are additional tests, such as run-tests, sub-grouped or CUSUM techniques, providing higher sensitivity to detect out-of-control states (Montgomery, 2007).

Observing the metric time series during the long monitoring period, they usually have out-of-control signals. Thus, in Figure 6, there are many observations below LCL or above UCL (green lines). It is simply due to the existence of special causes, which significantly impacts the metric. However, after the fitting model (7), the time series of residuals is within the control limits (Figure 12). It indicates that survived input features of the model (7) are the major contributors of specific period-to-period variations of the metric.

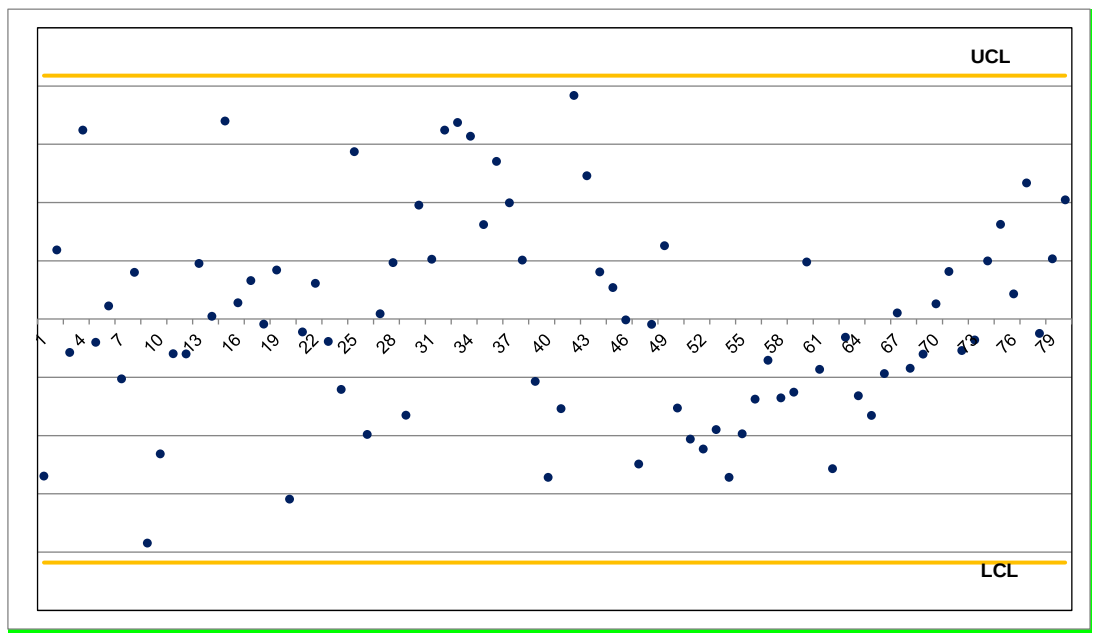

Figure 12. Time series of model residuals

SPC charts that are initially setup by in-control residual time series $R(t)$ can then be used for earlier detection of appearances of new contributing factors or changes in existing ones.

## 8. Forecasting, Stress Testing, and "What-If" Scenario Analysis

Formally, the trained model (7) can be used for forecasting, stress testing, and "What-If" scenario analysis providing future levels of the survived input features.

For example, forecasting the unemployment rate $X_1$ and assuming the number of accounts to be acquired per each of the segments $X_2$, while keeping all other factors unchanged, it is possible to predict future loss rates $\tilde{Z}$.

Alternatively, forecasted values of the survived input features can be replaced by stressed or assumed ("What-If") levels.

However, it should be noted that the approach is a backward-looking explanatory decomposition of the metric time series. It means that the final model (7), the survived input features, and their aggregated time series should be carefully evaluated with respect to forward-looking applications.

For example, for developed portfolios that have stable distributions of accounts along life stages, the tenure of account $X_3$ will normally be rejected in the final model even though it may be one of the most powerful predictors considering univariate model $G_3$. In this case, the trained model (7) will incorrectly predict no impact on the metric evaluating a new aggressive acquisition strategy.

## 9. Conclusions

Considering non-additivity of the metric, the large number of input features, limited monitoring periods, the dimensionality challenge of having two-dimensional arrays of input features and underlying variables, possible interactions or confounding of factors, and *a priory* unknown characters of the dynamic responses, it creates a very challenging analytical problem of interpreting changes of time series of the metric of interest.

The described approach consists of five basic steps: (1) segmentations along input features and the underlying variables calculating the metric of interest, (2) univariate or joint fittings of the calculated target metric by the aggregated input features on the segmented domains, (3) preliminary screening and transformations of input features according to the fitted models, (4) aggregation of the transformed features as time series, and (5) modelling the metric time series as combined effects of the aggregated features applying constrained linear regression.

These steps allow for an empirical backward-looking additive decomposition of the metric time series by time series of transformed input features. These features may include macroeconomic, seasonality, regulatory requirements, market conditions, acquisition characteristics, operational events, and managerial decisions.

Unsupervised variational autoencoders support both univariate and generalized segmentations when performing the first step of the approach.

Transformations based on univariate or joint models may include fitting non-linear functions with business-driven constraints (such as monotonicity). In addition, the transformation step may incorporate dynamic aspects of the metric responses, for example, a simple forward shift maximizing fitting of the model.

Existence of interactive effects should be noted as the concerning issue of the univariate approach. To mitigate this concern, the transformation can include interactions between input features as well. Practically, it makes sense to limit it to bivariate effects only. However, even a limited number of pairwise interactive effects can dramatically complicate interpretability.

The described approach can be applied to the entity distinct segments with sufficient sample sizes. It can provide better granularity on factor effects identifying sensitive and robust segments.

The additive form of the final model (7) allows for interpretation of the metric changes by contributing factors against the reference periods (15, 16).

In addition, individual contributions of entity elements can be learned by applying a trained model (8) to transformed input features.

When training univariate or joint models of transformations (4) for large entities (i.e., large $N(t)$), there is usually a low risk of overfitting. The major concern of overfitting is the last step of the approach of fitting model (7). Significant period-to-period variance of the metric and input features during the monitoring period impedes overfitting.

Statistical Process Control can be applied to residual time series. An "In-control" state means that there are no specific causes impacting the metric during the monitoring period other than the survived contributing factors in the fitted model (7).

Numerical differentiation (12) provides an alternative approach that allows for benchmarking obtained results followed by root-cause analysis if significant deviations are observed.

Formally, the trained model (7) can be used for forecasting, stress testing, and "What-If" scenario analysis providing future levels of the survived factors. However, it should be noted that the five-step approach is a backward-looking explanatory decomposition of the metric time series. It means that the final model (7), the survived input features, and their transformed time series should be carefully evaluated with respect to forward-looking applications.

Future research may include an application of graph neural networks for relational data (Schlichtkrull *et al*, 2017; Cvitkovic, 2020) considering the structure of relational datasets providing input features and underlying variables (Figure 1).

The discussed empirical decomposition can be seen as a research approach providing indicative insights on the historical variances of the non-additive metric of the entity by interpretable input features.

## Disclaimer

The paper represents the views of the author and do not necessarily reflect the views of the BMO Financial Group. All charts and data are simulated for illustrative purposes only and do not reflect the actual business state.